\documentclass{article}

\usepackage{arxiv}

\usepackage{graphicx}
\usepackage{xcolor}
\usepackage{caption}
\usepackage{subcaption}
\usepackage{amsmath}
\usepackage{amsfonts}
\usepackage[section]{placeins}
\usepackage{chngcntr}
\usepackage{soul}
\usepackage{ amssymb }
\usepackage[misc]{ifsym}
\usepackage{tikz}
\usepackage{pgfplots}
\usepgfplotslibrary{dateplot}
\usepgfplotslibrary{groupplots}
\usepackage{plotting}
\usepackage[ruled,linesnumbered,lined,commentsnumbered]{algorithm2e}
\usetikzlibrary{automata,positioning,arrows.meta,math,external}
\usetikzlibrary{decorations.pathreplacing}
\usetikzlibrary{shapes,shapes.geometric, snakes}
\usetikzlibrary{arrows, chains, fit, quotes}
\pgfplotsset{compat=1.16,
    tick label style = {font = {\fontsize{6pt}{12pt}\selectfont}},
    label style = {font = {\fontsize{8pt}{12pt}\selectfont}},
    legend style = {font = {\fontsize{8pt}{12pt}\selectfont}},
    title style = {font = {\fontsize{8pt}{12pt}\selectfont}}%,
  }

\newtheorem{Def}{Definition}

\usepackage[normalem]{ulem}
\usepackage{arxiv} 
\usepackage{natbib}

\begin{document}

\title{Multi-Objective Coordination Graphs for the \\ Expected Scalarised Returns with Generative Flow Models}

%\author{\name Conor F. Hayes$^{1,*}$, Timothy Verstraeten$^{2}$, Diederik M Roijers$^{2,3}$, Enda Howley$^{1}$, Patrick Mannion$^{1}$ \\\email $^{*}$c.hayes13@nuigalway.ie \\
%       \addr $^{1}$National University of Ireland Galway \\
%       $^{2}$Vrije Universiteit Brussel, Belgium \\
%       $^{3}$HU University of Applied Sciences Utrecht, the Netherlands
%       }

\author{Conor F. Hayes \\
       National University of Ireland Galway \\
       \texttt{c.hayes13@nuigalway.ie} 
       \AND
       Timothy Verstraeten \\
       Vrije Universiteit Brussel, Belgium \\
       \AND
       Diederik M. Roijers \\
       HU University of Applied Sciences Utrecht, the Netherlands\\
       Vrije Universiteit Brussel, Belgium 
       \AND
       Enda Howley \\
       National University of Ireland Galway \\
       \AND
       Patrick Mannion \\
       National University of Ireland Galway}

\maketitle

\begin{abstract}  % put your abstract here!
Many real-world problems contain both multiple objectives and agents, where a trade-off exists between objectives. Key to solving such problems is to exploit sparse dependency structures that exist between agents. For example in wind farm control a trade-off exists between maximising power and minimising stress on the systems components. Dependencies between turbines arise due to the wake effect. We model such sparse dependencies between agents as a multi-objective coordination graph (MO-CoG). In multi-objective reinforcement learning (MORL) a utility function is typically used to model a user's preferences over objectives. However, a utility function may be unknown a priori. In such settings a set of possibly optimal policies must be computed. Which policies are optimal depends on which optimality criterion applies. If the utility function of a user is derived from multiple executions of a policy, the scalarised expected returns (SER) must be optimised. If the utility of a user is derived from the single execution of a policy, the expected scalarised returns (ESR) criterion must be optimised. For example, wind farms are subjected to constraints and regulations that must be adhered to at all times, therefore the ESR criterion must be optimised. For MO-CoGs, the state-of-the-art algorithms can only compute a set of optimal policies for the SER criterion, leaving the ESR criterion understudied. To compute a set of optimal polices under the ESR criterion, also known as the \emph{ESR set}, distributions over the returns must be maintained. Therefore, to compute a set of optimal policies under the ESR criterion for MO-CoGs, we present a novel \emph{distributional multi-objective variable elimination} (DMOVE) algorithm. We evaluate DMOVE in realistic wind farm simulations. Given the returns in real-world wind farm settings are continuous, such as generated power, we utilise a generative flow model known as \emph{real-NVP} to learn the continuous return distributions to calculate the \emph{ESR set}.
\end{abstract}

\keywords{
multi-objective \and multi-agent \and coordination graphs \and (nonserial) dynamic programming \and expected scalarised returns \and generative flow models}

\maketitle

%%%%%%%%%%%%%%%%%%%%%%%%%%%%%%%%%%%%%%%%%%%%%%%%%%%%%%%%%%%%%%%%%%%%%%%%%%%%%%%%%%%%%%%%%%%%%%%%%%%%%%%%%
%% start of main body of paper

\section{Introduction}

%\timo{Given that the work introduces a new algorithm for more generic settings, and that wind farm control is just a realistic benchmark for it, I propose to remove the "wind farm control" part. I think the relevance of the algorithm is clear given that it is applied in a realistic setting, but adding it to the title seems to limit the scope unnecessarily. I.e., it is likely only wind farm control people will read the work, rather than the AI community, which still seems to be the target audience.}

%\dmr{Make sure to use the words non-serial dynamic programming (variable elimination), as dynamic programming seems to be one of the few key words in the CfP we can latch onto. Furthermore, make sure to state that we learn from interaction data (wind farm simulations) and then do our new non-serial dynamic programming algorithm, dmove. Also mention that we could in the future possibly integrate dmove into a online algorithms for multi-agent RL based on e.g. MAUCE and/or MATS, and  Pareto-UCB and/or Pareto Thompson Sampling. We need to make sure that some reviewer isn't just going to say: this is planning, and therefore out of scope of the workshop.}

In many real-world problems a trade-off must be made between multiple, often conflicting objectives \citep{vamplew2021scalar, reymond2022covid}. For example, learning a wind farm control policy is a multi-objective problem \citep{padullaparthi2022falcon}. Generally, a wind turbine aims to maximise its power production; however, this can lead to higher stress being induced on the turbine's components. Therefore, there is a trade-off between power production and induced damage. However, real-world problems can also have multiple agents \citep{radulescu2020survey, HayesTLO}. In this work, we focus on multi-objective multi-agent settings. 

Many multi-agent settings have a sparse dependency structure between the different agents, which can be exploited. For example, a phenomenon known as the wake effect occurs when upstream turbines extract energy from the wind, which increases the turbulence intensity of the wind flow and reduces the available energy for downstream turbines \citep{gonzalez2012wake}. In order to compensate for this effect, upstream turbines may aim to change their yaw angle, i.e., the direction that they are facing w.r.t. the incoming wind vector. This leads to a deflection of the wake, which reduces the power production of the misaligned turbines, but increases the energy production of the downstream turbines, which may lead to a higher energy production over the entire wind farm \citep{vandijk2016}. Therefore, it is crucial to consider the actions taken by different agents jointly to guarantee optimality. However, deriving a joint action control policy by directly considering the actions jointly is intractable given the number of possible solutions scales exponentially with the number of agents in the system. Fortunately, the dependency structure between agents is often sparse. In the case of the wind farm control setting, we can exploit the wake effect and the resulting dependencies, given that the decisions made by upstream turbines only affect neighboring downstream turbines \citep{verstraeten2020multi}. To this end, we aim to model a wind farm as a multi-objective coordination graph (MO-CoG) \citep{rollon2006bucket,marinescu2012multi, dellefave2011}.
%\dmr{cite a bunch? Rollon, Marinescu, Delle Fave, etc.?}.

In the multi-objective reinforcement learning (MORL) literature a utility function is typically used to model a user's preferences over objectives \citep{hayes2021practical}. In the real-world a user's utility function may be unknown. In this case, a set of optimal policies must be computed \citep{roijers2013survey}. In MORL multiple optimality criteria exist. Which criteria to optimise is determined by how the utility of a user is derived. If the utility of a user is derived from multiple executions of a policy \citep{roijers2013survey}, the scalarised expected returns (SER) criterion must be used to learn a set of optimal policies (for example the Pareto front \citep{Sebag2012}). However, if the utility of a user is derived from a single execution of a policy \citep{roijers2018multi, hayes2021dmcts}, then the expected scalarised returns (ESR) criterion should be optimised. The set of optimal policies under the ESR criterion is known as the \emph{ESR set} which must be computed using return distributions \citep{hayes2021esr_set, hayes2021expected}. It is important that the correct optimality criteria is selected, given the policies, and set of optimal policies, computed under the SER criterion and ESR criterion can be different \citep{Radulescu2019Equilibria, hayes2022modvi_esr, hayes2022decision}. For example, a wind farm controller is subjected to various restrictions, such as contractual constraints or grid regulations. It is important to adhere to these at all times. Therefore, the utility should be defined on the basis of a single policy execution, and thus the ESR criterion should be optimised. %\dmr{We don't actually take these constraints into account in our experiments as objectives, do we?}

Given return distributions must be utilised to compute the ESR set, a distributional approach to MORL must be taken. Recently, a number of distributional MORL algorithms, which utilise categorical distributions, have been deployed to learn policies \citep{hayes2022decision, reymond2021esractor, hayes2021dmcts}. In real-world problems the returns are often continuous. For example, the power and turbulence values for a wind turbine are continuous. While categorical distributions can be used, they cannot accurately and efficiently represent return distributions with continuous returns. Therefore, methods which can compute return distributions when the returns are continuous must be utilised in real-world settings like wind farm control.

The state-of-the-art MO-CoG algorithms cannot compute sets of optimal policies for the ESR criterion \citep{hayes2022decision, hayes2022modvi_esr}, because they use expected value vectors \citep{rollon2008multi,delle2011bounded,marinescu2012multi,roijers15computing}. Therefore, we propose a new nonserial dynamic programming \citep{rosenthal1977nonserial} algorithm for MO-CoGs that we call \emph{distributional multi-objective variable elimination} (DMOVE). DMOVE maintains set of return distributions to compute a set of optimal policies under the ESR criterion. Given the returns for wind farm control are continuous we utilise a generative flow model \citep{durkan2019spline, ho2019flow++} known as \emph{real-NVP} \citep{dinh2016density} to learn return distributions for each policy. We evaluate DMOVE using the state-of-the-art FLORIS wake simulator~\citep{FLORIS_2021} with different realistic wind farm configurations of varying sizes and dependencies. Specifically, we first learn a model of the vector-valued local reward functions in the MO-CoG from interacting with FLORIS, and then employ DMOVE to extract the \emph{ESR set}.

\section{Background}
In this section we introduce the relevant background information on multi-objective coordination graphs, dominance criteria in multi-objective reinforcement learning, and generative flow models.

\subsection{Multi-Objective Coordination Graphs}
In the context of multi-objective reinforcement learning (MORL) \citep{hayes2021practical, roijers2013survey}, it is possible to utilise multi-objective coordination graphs (MO-CoG) \citep{rollon2008multi} to compute a set of optimal policies. A MO-CoG is a multi-objective extension to coordination graphs and is a tuple $(\mathcal{D, \mathcal{A}, \mathcal{U}})$. $\mathcal{D} = \{1, ..., n\}$ is a set of $n$ agents. $\mathcal{D}$ is factorised into $p$, possibly overlapping, groups of agents $\mathcal{D}^{e}$. $\mathcal{A} = \mathcal{A}_{i}, ..., \mathcal{A}_{n}$ is the set of joint actions. $\mathcal{A}^{e}$ denotes the set of local joint actions for the group $\mathcal{D}^{e}$. $\mathcal{U} = {\textbf{u}^{1}, ..., \textbf{u}^{p}}$ is a set of $p$, $d$-dimensional local payoff functions. The joint payoff for all agents is the sum of local payoff functions: $\textbf{u}(\textbf{a}) = \sum_{e=1}^{p} \textbf{u}^{e}(\textbf{a}^{e})$.
The dependencies between the local reward functions and agents can be represented by a bipartite graph with a set of nodes $\mathcal{D}$ and a set of edges, $\mathcal{E}$. In this setting the nodes, $\mathcal{D}$, are agents and components of a factored payoff function, and an edge $(i, \textbf{u}^{e}) \in \mathcal{E}$ exists if and only if agent $i$ influences component $\textbf{u}^{e}$. The set of all possible join action value vectors is denoted by the set $\mathcal{V}$. 

%Given all agents share the payoff function $\textbf{u}(\textbf{a})$, a factor graph can be used to represent the decomposition of $\textbf{u}(\textbf{a})$ into local payoff functions. A factor graph has two types of vertices where agents are variables and local payoff functions are factors. Edges then connect local payoff function to the agents in their scope. 

Multi-objective variable elimination (MOVE) extends traditional variable elimination \citep{koller2009probabilistic} to problems with multiple-objectives. MOVE methods have been used to compute sets of optimal policies for MO-CoGs. However, the present MOVE methods compute solution sets for expected (or deterministic) reward vectors \citep{roijers15computing, rollon2006bucket} and do not take the underlying distributions over the rewards into account. Therefore, they are suitable for  the scalarised expected returns (SER) criterion, but not for the expected scalarised returns (ESR) criterion.

\subsection{Dominance Criteria in Multi-Objective Reinforcement Learning}
\label{sec:background_dom_criteria_modem}
In multi-objective reinforcement learning (MORL) \citep{hayes2021practical}, a user's preferences over objectives are represented by a utility function. When applying a user's utility function, the MORL literature generally focuses on two optimality criteria \citep{roijers2013survey}. Calculating the expected value of the return of a policy before applying the utility function leads to the scalarised expected returns (SER) optimisation criterion: 
\begin{equation}
    V_{u}^{\pi} = u\left(\mathbb{E} \left[ \sum\limits^\infty_{t=0} \gamma^t {\bf r}_t \:\middle|\: \pi, \mu_0 \right]\right).
    \label{eqn:ser}
\end{equation}
The SER criterion is the preferred optimality criteria is scenarios where the user aims to maximise their utility over multiple policies executions \citep{hayes2021practical}. The majority of MORL literature focuses on the SER criterion \citep{Sebag2012, white1982multi}.
Applying the utility function to the returns and then calculating the expected value leads to the ESR optimisation criterion:
\begin{equation}
    V_{u}^{\pi} = \mathbb{E} \left[ u\left( \sum\limits^\infty_{t=0} \gamma^t {\bf r}_t \right) \:\middle|\: \pi, \mu_0 \right].
    \label{eqn:esr}
\end{equation}
In certain settings a user's utility may be derived from a single execution of a policy, therefore the expected scalarised returns criterion should be optimised \citep{hayes2021expected, hayes2021dmcts, hayes2021dmcts_long, roijers2018multi,malerba2021esr}. 

In many scenarios the utility function of a user is unknown a priori, and set of optimal policies must be computed \citep{roijers2013survey}. To compute a set of optimal policies under the ESR criterion, a distribution over the returns must be maintained \citep{hayes2021esr_set, hayes2021expected, hayes2022modvi_esr}. By utilising the distribution over the returns a dominance criterion known as ESR dominance can be used to determine a partial ordering over policies \citep{hayes2021esr_set, hayes2021expected}. To calculate ESR dominance, the cumulative distribution function (CDF) of the each return distribution under consideration must be calculated. For a return distribution $\textbf{z}^{\pi}$, the CDF of $\textbf{z}^{\pi}$ is denoted by $F_{\textbf{z}^{\pi}}$. A return distribution $\textbf{z}^{\pi}$ ESR dominates a return distribution $\textbf{z}^{\pi'}$ if the following is true:
\begin{equation}
\textbf{z}^{\pi} >_{ESR} \textbf{z}^{\pi'} \Leftrightarrow \\  \forall \textbf{v}: F_{\textbf{z}^{\pi}}(\textbf{v}) \leq F_{\textbf{z}^{\pi'}}(\textbf{v})  \wedge \exists {\bf v} : F_{\textbf{z}^{\pi}}(\textbf{v}) < F_{\textbf{z}^{\pi'}}(\textbf{v}).
\label{eqn:esr_dominance}
\end{equation}
% If $\textbf{z}^{\pi}$ $>_{ESR}$ $\textbf{z}^{\pi'}$, then $\textbf{z}^{\pi}$ has a higher expected utility compared to $\textbf{z}^{\pi'}$ for all strictly monotonically increasing utility functions, $u$, 
%\begin{equation}
%    \textbf{z}^{\pi} >_{ESR} \textbf{z}^{\pi'} \Rightarrow \mathbb{E}(u(\textbf{z}^{\pi})) > \mathbb{E}(u(\textbf{z}^{\pi'})).
%\end{equation}

%The current state-of-the-art multi-policy MODeM methods focus almost exclusively on the SER criterion \citep{white1982multi, Sebag2012}, leaving the ESR criterion largely understudied \citep{hayes2021dmcts, hayes2021dmcts_long, malerba2021esr}. Given that the SER criterion and the ESR criterion utilise the utility function differently, SER methods cannot be used to compute a set of optimal policies for the ESR criterion. Additionally, a set of optimal policies under the SER criterion can exclude policies that are optimal under the ESR criterion \citep{hayes2021expected}. In all decision-making problems where a policy is only executed once, the ESR criterion must be utilised. As such problems are salient \citep{hayes2021practical}, new methods to compute a set of optimal policies for the ESR criterion must be developed to ensure optimal decision making in the real world.

\subsection{Generative Flow Models}
\label{sec:background_real_nvp}
To model return distributions where the returns are continuous we utilise generative flow models \citep{kobyzev2020normalizing, papamakarios2021normalizing} which are a class of probabilistic generative models \citep{tomczak2022deep} that learn non-linear models in continuous spaces through maximum likelihood. Generative flow models utilise bijective functions which enable exact and tractable density evaluation and inference on continuous data using the change of variables formula \citep{kingma2018glow}. 

Consider an observed data point $x$ generated from sampling a random variable $X$, a latent variable $Z$ and a bijective function $f : X \rightarrow Z$ (and thus $f^{-1} : Z \rightarrow X$ exists). Therefore, the change of variables formula can be used to define a model distribution on the random variable $X$ as follows:
\begin{equation}
    p_{X}(x) = p_{Z}(f(x))\:\left|\textit{det} \left(\triangledown f(x)\right)\right|,
\end{equation}
where $\triangledown f(x)$ is the Jacobian of $f$ at $x$. To generate samples from the original sample space, $X$, a random sample, $z$, is drawn from the latent space, $Z$, according to $p_{Z}$, and the inverse, $x = f^{-1}(z)$ is computed. In this work we use a model known as \emph{real-valued non-volume preserving transformations} (real-NVP) \citep{dinh2016density} which utilises the fact the determinant of a triangular matrix can be efficiently computed as the product of its diagonal terms to learn a tractable and flexible model. 

\citet{dinh2016density} stack a sequence of simple bijections, where each simple bijection is updated using only part of the input vector with a function which is simple to invert. \citet{dinh2016density} refer to each simple bijection in the sequence as an \emph{affine coupling layer}. Each \emph{affine coupling layer} takes a $D$ dimensional input $x$ and $d$, where $d < D$ and outputs $y$. The output $y$ can be defined as follows:
\begin{equation}
    y_{1:d} = x_{1:d}
\end{equation}
\begin{equation}
    y_{(d+1):D} = x_{(d+1):D} \odot \exp\left(s(x_{1:d})) + t(x_{1:d}\right),
\end{equation}
where $s$ is a scale function, $t$ transformation function and are functions from $\mathcal{R}^{d} \rightarrow \mathcal{R}^{D-d}$ and $\odot$ is the element-wise product. \citet{dinh2016density} show that the Jacobian of the transformation is triangular and therefore it is efficient to compute the determinant. A limitation of utilising \emph{affine coupling layers} is during the forward transformation some components are left unchanged. \citet{dinh2016density} overcome this issue by combining \emph{affine coupling layers} in an alternating pattern, therefore elements which are left unchanged in one transformation are modified in the next (see supplementary information (SI) for more details). %\dmr{I think Appendix C has very little text right now, and could do with a bit more explanation. Also, Shouldn't we say in the main text that what is described in appendix C is \emph{our} configuration?}
%\conor{I was pointing to the wrong part of the appendix, updated now.}

\section{Multi-Objective Coordination Graphs for the Expected Scalarised Returns}
The current literature on multi-objective coordination graphs (MO-CoGs) focuses exclusively on the SER criterion \citep{rollon2006bucket, roijers15computing}. As shown by \citet{hayes2022modvi_esr}, methods that compute solution sets for the SER criterion cannot be used under the ESR criterion. Before we outline our new method to compute solution sets for the ESR criterion, we must define a set of optimal policies under the ESR criterion for MO-CoGs.

To determine a partial ordering over policies under the ESR criterion, a dominance relation known as ESR dominance can be used \citep{hayes2021expected}. To calculate ESR dominance a return distribution for each local payoff function must be calculated. For MO-CoGs we define a return distribution, $\textbf{z}$, as the distribution over the returns of a local payoff function. Therefore, we define $\mathcal{Z}$ where $\mathcal{Z} = {\textbf{z}^{1}, ..., \textbf{z}^{p}}$ is a set of $p$, $d$-dimensional return distributions of local payoff functions. The joint payoff for all agents is the sum of local payoff return distributions: $\textbf{z}(\textbf{a}) = \sum_{e=1}^{p} \textbf{z}^{e}(\textbf{a}^{e})$. The set of all possible joint action return distributions is denoted by $\mathcal{V}$.

By utilising ESR dominance (Equation \ref{eqn:esr_dominance}) we can define a set of optimal policies under the ESR criterion for a MO-CoG as a set of ESR non-dominated global joint actions $\textbf{a}$ and associated return distributions of local payoff functions, $\textbf{z}(\textbf{a})$, known as the $\emph{ESR set}$:
\begin{Def}
The ESR set (ESR) of a MO-CoG, is the set of all joint actions and
associated payoff return distributions that are ESR non-dominated,
\begin{equation}
    ESR(\mathcal{V}) = \left\{\textbf{z}(\textbf{a}) \in \mathcal{V}\  \middle|\ \nexists\ \textbf{z}(\textbf{a})^{'} \in \mathcal{V} : \textbf{z}(\textbf{a})^{'} \succ_{\text{ESR}}\textbf{z}(\textbf{a})\right\}.
\label{eqn:esr_set}
\end{equation}
\end{Def}

\section{Distributional Multi-Objective Variable Elimination}
To solve multi-objective coordination graphs (MO-CoGs) for the expected scalarised returns (ESR) criterion we define a novel \emph{distributional multi-objective variable elimination} (DMOVE) algorithm. DMOVE utilises return distributions and ESR dominance to compute the \emph{ESR set}.

Generally multi-objective variable elimination (MOVE) methods translate the problem to a set of value set factors \citep{roijers2013survey, roijers15computing}. Given under the ESR criterion return distributions must be utilised, the problem must first be translated to a set of return distribution set factors (RSFs), $f$, where each RSF $f^{e}$ is a function mapping local joint actions to set of payoff return distributions. On initialisation each RSF is a singleton set containing a local actions payoff return distribution and is defined as follows:
 \begin{equation}
     f^{e}(\textbf{a}^{e}) = \{ \textbf{z}^{e} (\textbf{a}^{e}) \}
     \label{eqn:rsf}
 \end{equation}
It is possible to describe the coordination graph as a bipartite graph whose nodes, $\mathcal{D}$, are both agents and components of a factored RSF, and an edge $(i, f^{e}) \in \mathcal{E}$ if and only if agent $i$ influences component $f^{e}$. It is important to note an agent node is joined by an edge to a factored RSF component if the agent influences the RSF. Therefore, the dependencies can be described by setting $\mathcal{E} = \{ (i, f^{e}) | i \in \mathcal{D}^{e}\}$.
%and exploits the transitive properties of ESR dominance.
To compute an \emph{ESR set}, DMOVE  treats a MO-CoG as a series of local sub-problems. DMOVE manipulates the set of RSFs by computing local \emph{ESR sets} (LESRs) when eliminating agents. To calculate LESRs we first must define a set of neighbouring RSFs, $f_{i}$. 
\begin{Def}
The set of neighbouring RSFs $f_i$ of agent $i$ is the subset of all RSFs which agent $i$ influences.
\end{Def}
%In other words, the set of neighbouring RSFs $f_i$ of agent $i$ is the subset of all RSFs which agent $i$ connects to via an edge. 
Each agent $i$ has a set of neighbour agents, $n_{i}$, where each agent in $n_{i}$ influences one or more RSFs in $f_{i}$ (see SI for a visualisation). To compute a LESR, all return distributions for the sub-problem must first be considered, $\mathcal{V}_{i}$, by calculating the following:
\begin{equation}
    \mathcal{V}_{i}(f_{i}, \textbf{a}^{n_{i}}) = \bigcup_{a^{i}} \bigoplus_{f^{e} \in f_{i}} f^{e}(\textbf{a}^{e}),
    \label{eqn:subproblem_sum}
\end{equation}
where $\odot$ is the cross sum of sets of return distributions. In Equation \ref{eqn:subproblem_sum} all actions in $\textbf{a}^{n_{i}}$ are fixed however, not $a^{i}$, and $\textbf{a}^{e}$ is formed from $a^{i}$ and the appropriate part of $\textbf{a}^{n_{i}}$. Once $\mathcal{V}_{i}(f_{i}, \textbf{a}^{n_{i}})$ has been computed for agent $i$, a LESR can be calculated by applying an ESR pruning operator. Therefore, a LESR is defined as follows:
\begin{Def}
A local ESR set, a LESR, is the ESR non-dominated subset of $\ \mathcal{V}_{i}(f_{i}, \textbf{a}^{n_{i}})$:
\begin{equation}
    LESR_{i}(f_{i}, \textbf{a}^{n_{i}}) =  ESR \ (\ \mathcal{V}_{i}(f_{i}, \textbf{a}^{n_{i}})\ ),
    \label{eqn:lesr}
\end{equation}
\end{Def}
When calculating a LESR a new RSF, $f_{new}$, is generated which is conditioned on the actions of the agents in $n_{i}$:
\begin{equation}
    \forall \ \textbf{a}^{n_{i}} \ f_{new}(\textbf{a}^{n_{i}}) = LESR_{i}(f_{i}, \textbf{a}^{n_{i}}).
    \label{eqn:f_new}
\end{equation}
The set of RSFs, $f$, must then be updated with $f_{new}$. Therefore, we remove the RSFs in $f_{i}$ from $f$ and update $f$ with $f_{new}$. To do so, we define a new operator, known as the \texttt{eliminate} operator:
\begin{equation}
    \texttt{eliminate}(f, i) = (f \setminus f_{i}) \cup \{ f_{new}(\textbf{a}^{n_{i}}) \}.
    \label{eqn:eliminate}
\end{equation}
Computing Equation \ref{eqn:f_new} and Equation \ref{eqn:eliminate} removes agent $i$ from the coordination graph. Therefore, the nodes and edges of the coordination graph are updated, where the edges for each agent in $n_{i}$ are now connected to the new RSF, $f_{new}$. 

Utilising the steps outlined above, we can now define the DMOVE algorithm (Algorithm \ref{alg:DMOVE}). DMOVE first translates the problem into a set of return distribution set factors (RSFs) and removes agents in a predefined order, \textbf{q}. DMOVE calls the algorithm $\texttt{eliminate}$ which computes local \emph{ESR sets} by pruning, and updates the set of RSFs. Once all agents have been eliminated the final factor from the set of RSFs, $f$, is retrieved, pruned and returned. The resulting set, $\mathcal{S}$, contains ESR non-dominated return distributions, known as the \emph{ESR set}, and the associated joint actions. DMOVE only executes a forward pass, and calculates joint actions using a tagging scheme \citep{roijers15computing}.

\begin{algorithm}
\SetAlgoLined
\textbf{Input}: $\mathcal{U}\ \leftarrow$ \text{A set of local payoff functions;}
$\textbf{q}\ \leftarrow \text{an elimination order}$ \\
$f \rightarrow$ \textit{create one RSF for every local payoff function in } $\mathcal{U}$ \\
\While{$\textbf{a}^{n_{i}} \in \mathcal{A}^{n_{i}}$}
{$i$ $\leftarrow$ $\texttt{q.dequeue()}$ \\
$f$ $\leftarrow$ $\texttt{eliminate}(f, i, \texttt{prune1, prune2})$
} 
$f$ $\leftarrow$ $\text{retrieve final factor from}$ $f$ \\
$\mathcal{S}$ $\leftarrow$ \texttt{prune3}($f(a_{\emptyset})$) \\
\textbf{Return $\mathcal{S}$}
 \caption{\texttt{DMOVE} ($\mathcal{U}$, \texttt{prune1}, \texttt{prune2}, \texttt{prune3}, \texttt{q})}
 \label{alg:DMOVE}
\end{algorithm}

The \texttt{eliminate} algorithm (Algorithm \ref{alg:eliminate}) calculates the LESRs and updates the set of RSFs, $f$. To calculate the LESRs the function \texttt{CalculateLESR} is utilised. We define $\texttt{CalculateLESR}_{i}$ as follows:
\begin{equation}
\texttt{CalculateLESR}_{i}(f_{i}, \textbf{a}^{n_{i}}, \texttt{prune1,prune2}) = \texttt{prune2}\left(\bigcup_{a^{i}} \bigoplus^{*}_{f^{e} \in f_{i}} f^{e}(\textbf{a}^{e})\right) ,
\end{equation}
where $\bigoplus\limits^{*}$ is the prune and cross-sum operator defined by \citet{roijers15computing}. The $\texttt{CalculateLESR}_{i}$ function prunes at two different stages; \texttt{prune1} is applied after the cross-sum has been computed and \texttt{prune2} is applied after the union over all $a^{i}$, this results in incremental pruning \citep{cassandra1997pruning}. 
\begin{algorithm}
\SetAlgoLined
\textbf{Input}: $f\ \leftarrow$ \text{A set of RSFs;} $i\ \leftarrow$ \text{an agent} \\
$n_{i}\ \leftarrow \text{the set of neighbouring agents of $i$;}$
$f_{i} \leftarrow$ \text{the subset of f that $i$ influences;}
$f_{new}(a^{n_{i}}) \leftarrow$ \text{a new RSF} \\
\For{$\textbf{a}^{n_{i}} \in \mathcal{A}^{n_{i}}$}
{$f_{new}(a^{n_{i}}) \leftarrow$ $\texttt{CalculateLESR}_{i}(f_{i}, a^{n_{i}}, \texttt{prune1,prune2})$} 
$f \leftarrow \ f \setminus \ f_{i} \cup \{ f^{new} \}$ \\
\textbf{Return $f$}
 \caption{\texttt{eliminate}($f$, $i$, \texttt{prune1}, \texttt{prune2})}
 \label{alg:eliminate}
\end{algorithm}
Both DMOVE (Algorithm \ref{alg:DMOVE}) and \texttt{eliminate} (Algorithm \ref{alg:eliminate}) are paramaterised by pruning operators. To compute the \emph{ESR set}, the pruning algorithm \texttt{ESRPrune} (see SI, \cite{hayes2022modvi_esr, hayes2022decision}) is used as each pruning operator.

The state-of-the-art MORL algorithms that compute sets of optimal solutions for the ESR criterion focus on utilising categorical distributions to represent return distributions \citep{hayes2022modvi_esr, hayes2021expected}. However, categorical distributions are discretised and therefore cannot fully represent the distribution of observations for continuous data. Therefore, we utilise a generative flow model \citep{tomczak2022deep} known as \emph{real non-volume preserving transformations} (\emph{real-NVP, \cite{dinh2016density}}, see Section \ref{sec:background_real_nvp}), to model return distributions with continuous returns. 

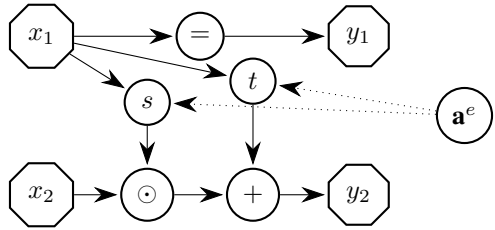
\begin{figure}[h]
    \centering
     \begin{tikzpicture}[>={Stealth[width=6pt,length=9pt]}, skip/.style={draw=none}, shorten >=1pt, accepting/.style={inner sep=1pt}, auto]
     \draw (0.0pt, 0.0pt)node[regular polygon,regular polygon sides=8, fill=none, thick, minimum height=0.45cm,minimum width=0.45cm, draw](0){$x_{1}$};
     \draw (60.0pt, 0.0pt)node[circle, thick, minimum height=0.6cm,minimum width=0.6cm, draw](1){$=$};
     \draw (120.0pt, 0.0pt)node[regular polygon,regular polygon sides=8, fill=none, thick, minimum height=0.6cm,minimum width=0.6cm, draw](2){$y_{1}$};
     \draw (0.0pt, -60.0pt)node[regular polygon,regular polygon sides=8, fill=none, thick, minimum height=0.6cm,minimum width=0.6cm, draw](3){$x_{2}$};
     \draw (40.0pt, -60.0pt)node[circle, fill=none, thick, minimum height=0.6cm,minimum width=0.6cm, draw](4){$\odot$};
     \draw (80.0pt, -60.0pt)node[circle, thick, minimum height=0.6cm,minimum width=0.6cm, draw](5){$+$};
     \draw (120.0pt, -60.0pt)node[regular polygon,regular polygon sides=8, fill=none, thick, minimum height=0.6cm,minimum width=0.6cm, draw](6){$y_{2}$};
     \draw (160.0pt, -30.0pt)node[circle, thick, minimum height=0.6cm,minimum width=0.6cm, draw](7){$\textbf{a}^{e}$};
     \draw (40.0pt, -25.0pt)node[circle, thick, minimum height=0.6cm,minimum width=0.6cm, draw](8){$s$};
     \draw (80.0pt, -17.0pt)node[circle, thick, minimum height=0.6cm,minimum width=0.6cm, draw](9){$t$};
     
     %\path[->] (6) edge[dotted, bend right = 50]
     \path[->] (0) edge[] node[above]{} (1);
     \path[->] (1) edge[] node[above]{} (2);
     \path[->] (7) edge[dotted] node[above]{} (8);
     \path[->] (7) edge[dotted] node[above]{} (9);
     \path[->] (8) edge[] node[above]{} (4);
     \path[->] (9) edge[] node[above]{} (5);
     \path[->] (3) edge[] node[above]{} (4);
     \path[->] (4) edge[] node[above]{} (5);
     \path[->] (5) edge[] node[above]{} (6);
     \path[->] (0) edge[] node[above]{} (8);
     \path[->] (0) edge[] node[above]{} (9);
     ;
      
    \end{tikzpicture}
    
    \caption{Computational graph for the forward propagation of a \emph{real-NVP} instance, conditioned on a local action, $\textbf{a}^{e}$.
    }
    \label{fig:cond_real_nvp_graph}
\end{figure}

As already mentioned to compute DMOVE, the problem must first be decomposed into a set of RSFs, $f$. To calculate $f$ a set of \emph{real-NVP} models, $\mathcal{X}$ is used, where a \emph{real-NVP} model exists per payoff function, $x^{e} \in \mathcal{X}$. To compute a return distribution for each local action, $\textbf{a}^{e}$, for each payoff function each \emph{real-NVP} model is conditioned on the appropriate local action $\textbf{a}^{e}$ during training. The local action, $\textbf{a}^{e}$, is then used to condition the scale, $s$, and transformation, $t$, functions for each individual \emph{affine coupling layer} as follows:
\begin{equation}
    y_{1:d} = x_{1:d}
\end{equation}
\begin{equation}
    y_{(d+1):D} = x_{(d+1):D} \odot exp(s(x_{1:d}, \textbf{a}^{e})) + t(x_{1:d}, \textbf{a}^{e}),
\end{equation}
where both $s$ and $t$ are represented by a neural network. Figure \ref{fig:cond_real_nvp_graph} displays a computational graph of the forward propagation of a conditioned \emph{real-NVP} instance for a single \emph{affine coupling layer}. As previously highlighted in Section \ref{sec:background_real_nvp}, during the forward transformation some components are left unchanged, therefore each \emph{real-NVP} model must have multiple \emph{affine coupling layers} to ensure each component is modified during the training phase.
To train a \emph{real-NVP} instance, samples from the underlying return distributions for each payoff function must be gathered. To do so, we call an algorithm \texttt{learn} which randomly executes a global joint action, $\textbf{a}$, in the environment. Once $\textbf{a}$ has been executed a set of vectorial returns $\textbf{r}$, is received and is decomposed into local returns, $\textbf{r}^{e}$, for each group. Each local return, $\textbf{r}^{e}$, and joint local action, $\textbf{a}^{e}$ are then pushed to a replay buffer, where each \emph{real-NVP} instance has its own replay buffer. Using the data stored in the replay buffer, it is possible to train each \emph{real-NVP} at different increments, $t_{inc}$, using a number of random samples from the replay buffer.

Once each \emph{real-NVP} model has been trained the set of RSFs $f$ can be calculated. To generate the required return distributions for each local action, $\textbf{a}^{e}$, for each payoff function, $\textbf{u}^{e}$, $\textbf{a}^{e}$ is passed as input to the corresponding \emph{real-NVP} instance, $x^{e}$. A random sample is taken from the latent space of $x^{e}$ and the inverse is then computed which generates an observation from the observed data space. To generate a return distribution for $\textbf{a}^{e}$ this process is repeated $n$ times. The constructed return distribution for each local action, $\textbf{a}^{e}$, can be used to construct the singleton set outlined in Equation \ref{eqn:rsf} for each RSF in $f$.  

The following algorithm \texttt{learn} (Algorithm \ref{alg:learn}) is defined which outlines how each \emph{real-NVP} instance is trained and calls DMOVE, which returns a set of optimal return distributions and corresponding global joint actions.

\begin{algorithm}
\SetAlgoLined
\textbf{Input}: $steps$ \ $\leftarrow$ $\text{a number of random global joint actions to execute;}$ $\mathcal{U}$ $\leftarrow$ $\text{a set of payoff functions}$ \\ $\mathcal{X}$ \ $\leftarrow$ \ $\emptyset ;$
$\mathcal{F}$ \ $\leftarrow$ \ $\emptyset ;$ $t_{inc} \leftarrow \text{increments for real-NVP training}$ \\
\For{$u$ $\in$ $\mathcal{U}$}{$\mathcal{X}$ $\texttt{.append(real-NVP())}$}
\For{$step \in steps$}
{
$\textbf{a}$ $\leftarrow$ $\text{random global joint action}$ \\
$\textbf{r}$ = $\text{env.execute(}\textbf{a}\texttt{)}$\\
\For{$x^{e}$ $\in$ $\mathcal{X}$}
{
$x^{e}$$\text{.push(}$$\textbf{r}^{e}, \textbf{a}^{e}$$\text{)}$ \\
\If{$step = t_{inc}$}
{
$x^{e}$$\text{.train()}$
}
}
}
$\mathcal{S}$ $\leftarrow$ $\text{DMOVE()}$
\caption{\texttt{learn()}}
\label{alg:learn}
\end{algorithm}

\section{Experiments}
Generally, offshore wind farms are designed in a symmetric grid-like structure, given this approach is advantageous for planning and construction of the wind farms \citep{tao2020topology}. Utilising a grid-like structure can cause wake given the turbines in the farm are in close proximity \citep{tao2020topology}. To construct the dependencies required to model the problem as a MO-CoG, we utilise a geometric approach to analyse the wake field for a given wind direction as proposed by \citet{verstraeten2021scalable}. To do so, a downstream turbine, $t$, is considered to be dependent on a reference turbine, $t_{ref}$ if and only if $t$ is geographically positioned within a specified radius and is within a specified angle from the incoming wind vector. We consider that downstream turbines located at an angle of $22.5^\circ$ and a radius of $1$km apart from a given reference turbine lie outside the wake field of that reference turbine \citep{verstraeten2021scalable}. Using the identified dependencies it is possible to model the problem as a MO-CoG. We deploy DMOVE to compute the \emph{ESR set} using several wind farms with different numbers of turbines and different grid-like structures.

\begin{figure}[h]
        \centering
        \begin{subfigure}[b]{0.2\textwidth}
            \centering
            \includegraphics[height=85pt,width=100pt]{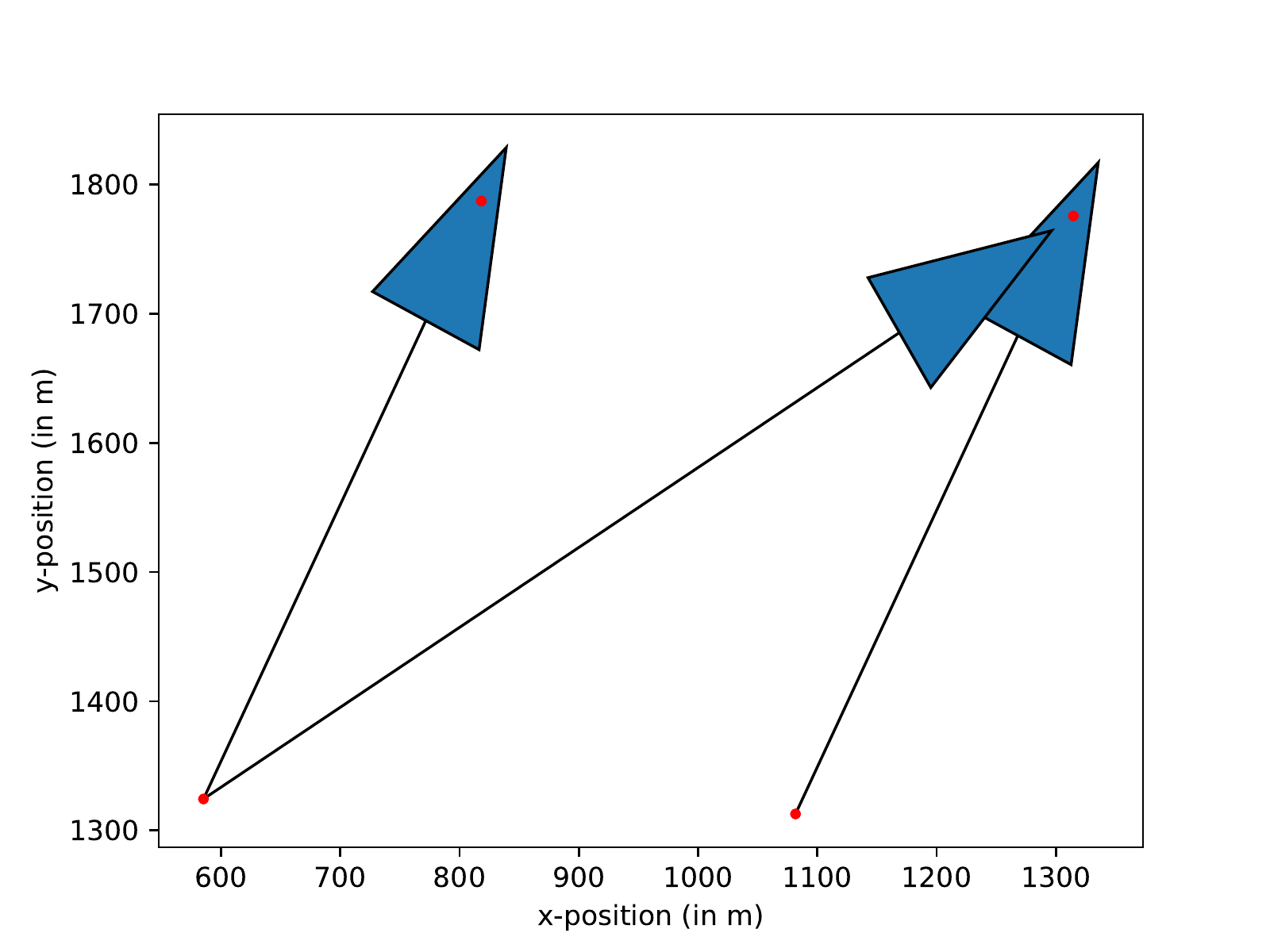}
        \end{subfigure}
        \begin{subfigure}[b]{0.2\textwidth}
            \centering
            \includegraphics[height=85pt,width=100pt]{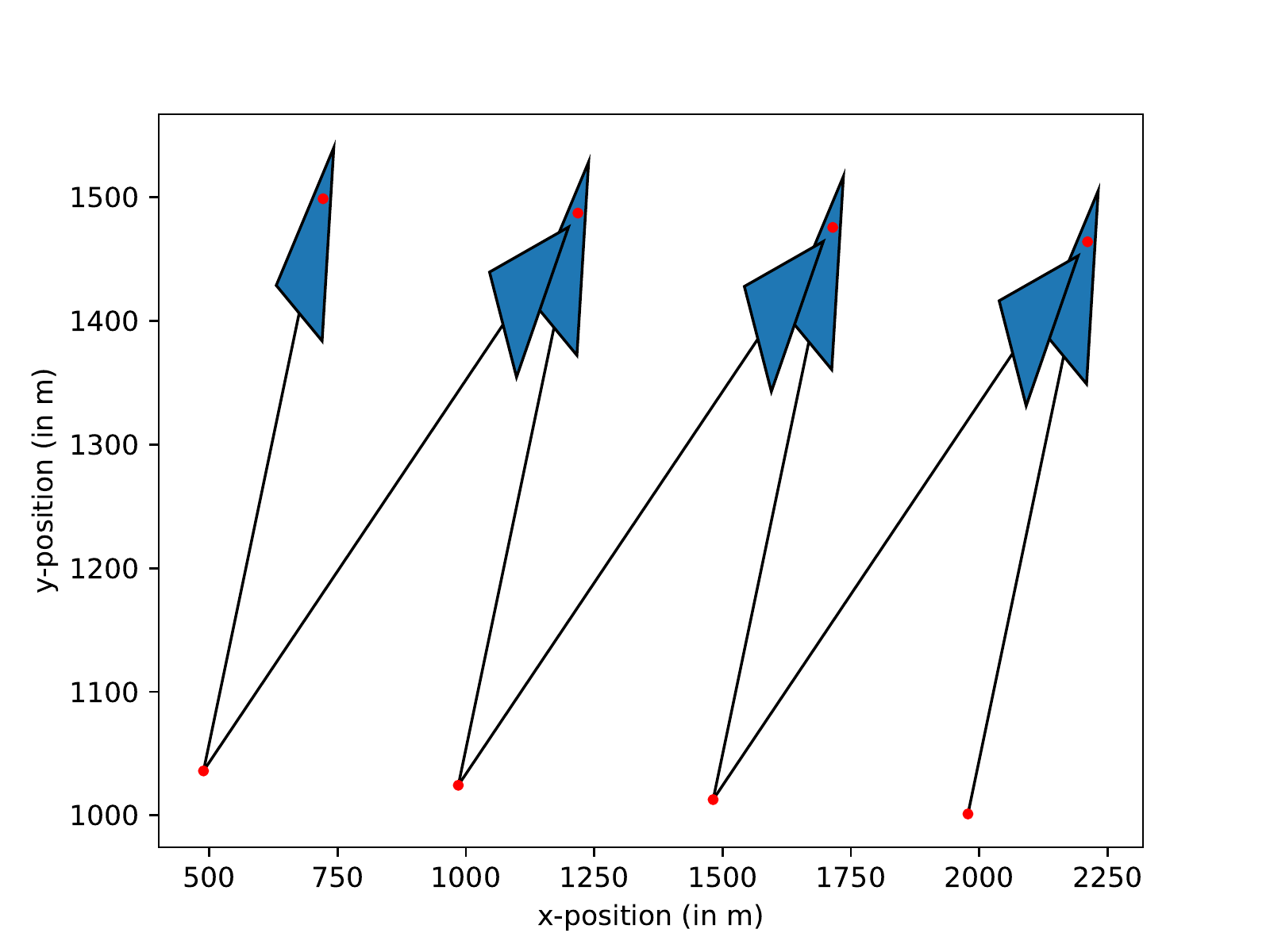}
        \end{subfigure}
        \begin{subfigure}[b]{0.2\textwidth}
            \centering
            \includegraphics[height=85pt,width=100pt]{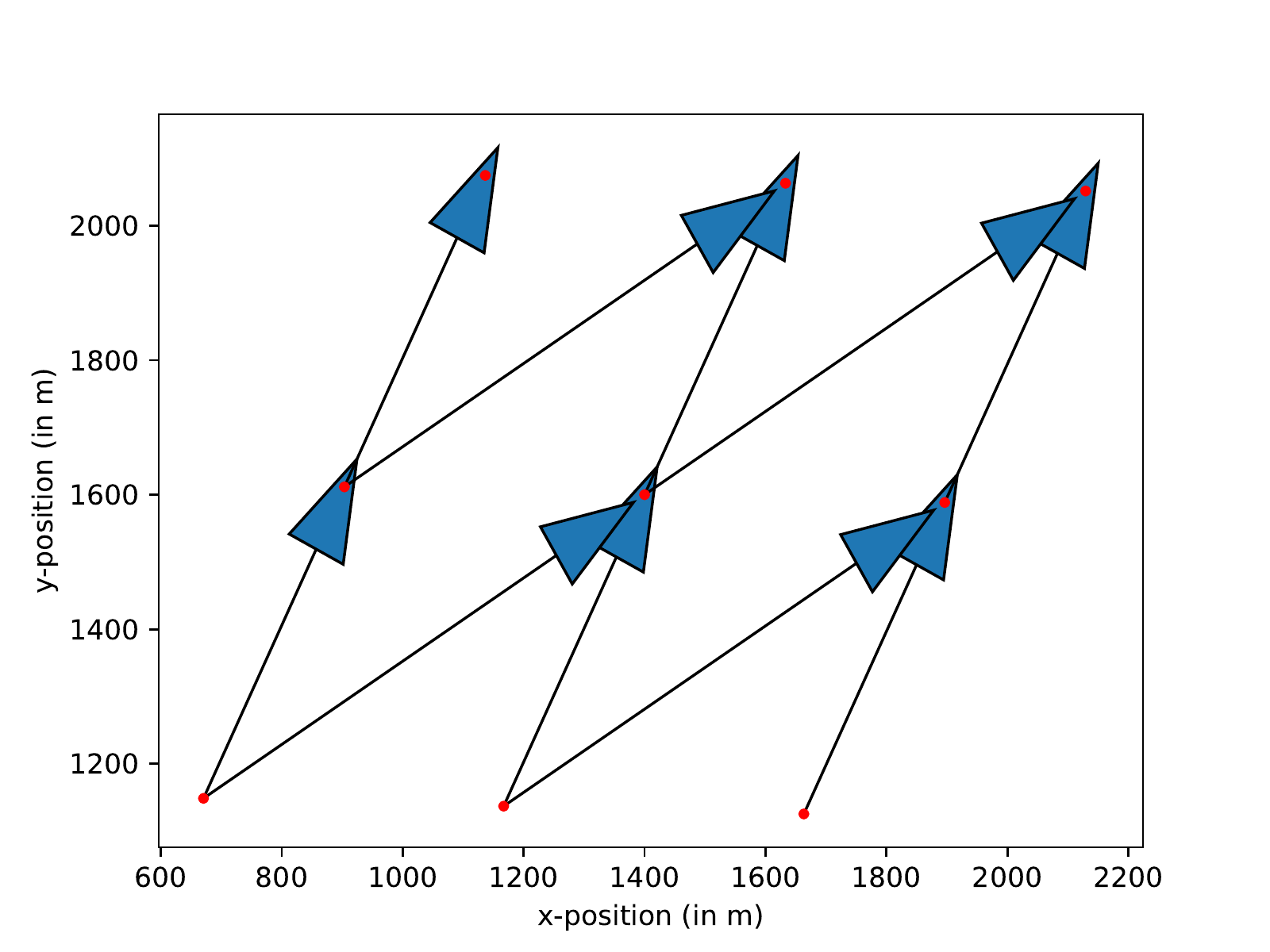}
 
        \end{subfigure}
        \begin{subfigure}[b]{0.2\textwidth}
            \centering
            \includegraphics[height=85pt,width=100pt]{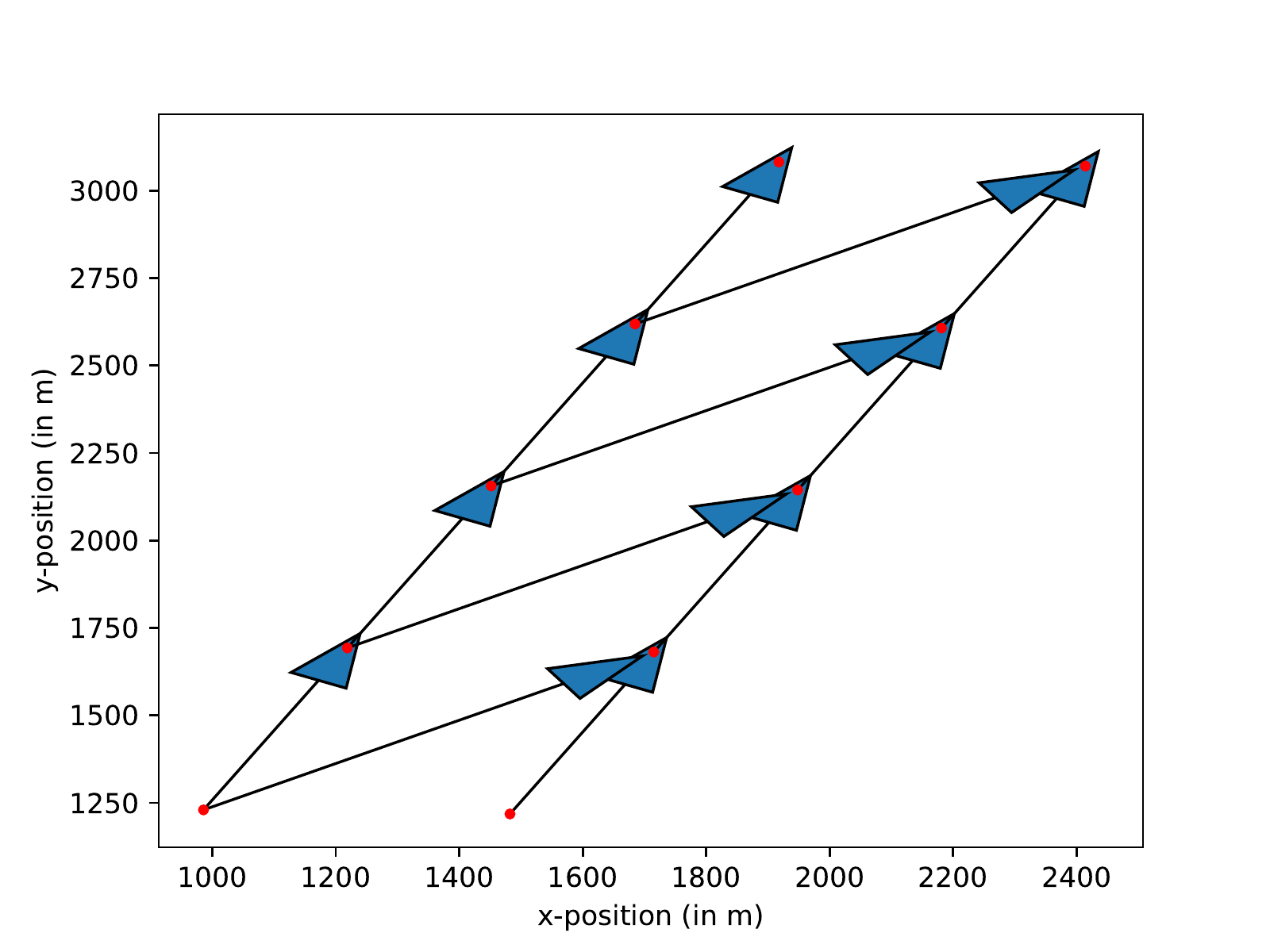}

        \end{subfigure}
        \caption{The dependency graphs for wind farms with 4, 8, 9 and 10 turbines.}\label{fig:dependency_graphs}
\end{figure}

Each wind farm has specific wind conditions. For each wind farm we investigate the incoming wind vector at $30^\circ$ with a wind speed of $11$ m/s \citep{verstraeten2021scalable}. Under the conditions presented the wake effect is strong and the dependencies for each farm created by the wake effect are shown in Figure \ref{fig:dependency_graphs}. For each wind farm, each turbine is placed $500$m apart along the x-axis and $400$m apart on the y-axis \citep{verstraeten2021scalable}. 

%Developed by the National Renewable Energy Lab in the USA, FLORIS can simulate the wind farm and associated wake conditions. 
To simulate each wind farm at the specified conditions, a state-of-the-art simulator known as FLORIS is used \citep{FLORIS_2021}. A reference turbine model for the LW-8MW turbine is also used to model the turbines dynamics \citep{desmond2016description}. These models can accurately represent the scale and conditions of offshore wind farms.

We aim to use DMOVE to compute a set of optimal policies under the ESR criterion where the turbulence intensity and power for each turbine is maximised by optimising the yaw for each turbine. We choose to minimize turbulence intensity, as it is a direct contributor to fatigue loads, and thus to the overall lifetime of the turbine \citep{frandsen1999integrated}. DMOVE considers each turbine as an agent. Each turbine has the following possible yaw angles, $y$:
\begin{equation}
    y = [-10, -5, \ 0, \ 5, \ 10]
\end{equation}
Therefore, the possible yaw angles can be considered as the action space for each agent \citep{verstraeten2020multi}. For each joint action, \textbf{a}, a reward $\textbf{r}$ is received where the reward can be decomposed into reward vectors for each group, $\textbf{r}^e$, as follows:
\begin{equation}
    \textbf{r}^e = [-\text{turbulence}, \ \text{power}].
\end{equation}
Therefore, DMOVE aims to find a set of optimal return distributions over the turbulence and power returns, and the associated global yaw angles. All required parameters for DMOVE are presented in SI.

\begin{figure}[h]
    \centering
    
\begin{tikzpicture}
    \begin{groupplot}[
    group style={
            % set how the plots should be organized
            group size=4 by 1,
            % only show ticklabels and axis labels on the bottom
            %y labels at= edge left,
            x descriptions at=edge bottom,
            % set the `vertical sep' to zero
            horizontal sep=30pt,
            vertical sep=0pt
        },
      height = 3.25cm,
      width=4.5cm]%,
      %axis y line*=left,
      %xlabel={power},
      %ylabel={turbulence},
    ]
    
    \nextgroupplot[ylabel={turbulence}, xlabel = {power}]
      \addplot[only marks,draw opacity=0.5,mark=x,blue] table [x=power, y=turbulence, col sep=comma] {figures/mo_floris/4_turbines/esr_set/esr_set.csv};
    \nextgroupplot[xlabel = {power}]
      \addplot[only marks,draw opacity=0.5,mark=x,blue] table [x=power, y=turbulence, col sep=comma] {figures/mo_floris/8_turbines/esr_set/esr_set.csv};
  \nextgroupplot[xlabel = {power}]
      \addplot[only marks,draw opacity=0.5,mark=x,blue] table [x=power, y=turbulence, col sep=comma] {figures/mo_floris/9_turbines/esr_set/esr_set.csv};
    \nextgroupplot[xlabel = {power}]
      \addplot[only marks,draw opacity=0.5,mark=x,blue] table [x=power, y=turbulence, col sep=comma] {figures/mo_floris/10_turbines/esr_set/esr_set.csv};
    \end{groupplot}
  \end{tikzpicture}
  
  \caption{The expected value vector for each policy in \emph{ESR set} for wind farms with 4, 8, 9 and 10 turbines.}
    \label{fig:esr_set}
\end{figure}
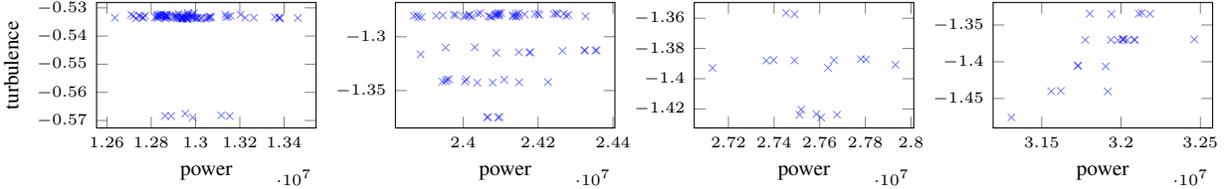
\begin{figure}[h]
    \centering
    
\begin{tikzpicture}
    \begin{groupplot}[
    group style={
            % set how the plots should be organized
            group size=4 by 1,
            % only show ticklabels and axis labels on the bottom
            %y labels at= edge left,
            x descriptions at=edge bottom,
            % set the `vertical sep' to zero
            horizontal sep=30pt,
            vertical sep=15pt
        },
      height = 3.25cm,
      width=4.5cm]%,
      %axis y line*=left,
      %xlabel={power},
      %ylabel={turbulence},
    ]
    
    \nextgroupplot[xlabel = {power}, grid=major,ymin=-0.101,ymax=-0.096, ylabel={turbulence}, yticklabel style={/pgf/number format/.cd,fixed,precision=3}]
      \addplot[only marks, draw opacity=0.5,mark=x,red] table [x=learned_dist_power, y=learned_dist_turbulence, col sep=comma] {figures/mo_floris/4_turbines/data/dist_1.csv};
      \addplot[only marks,draw opacity=0.5,mark=x,blue] table [x=sampled_dist_power
, y=sampled_dist_turbulence, col sep=comma] {figures/mo_floris/4_turbines/data/dist_1.csv};

    \nextgroupplot[xlabel = {power}, grid=major,ymin=-0.205,ymax=-0.197, yticklabel style={/pgf/number format/.cd,fixed,precision=3}]
      \addplot[only marks,draw opacity=0.5,mark=x,red] table [x=learned_dist_power, y=learned_dist_turbulence, col sep=comma] {figures/mo_floris/4_turbines/data/dist_2.csv};
      \addplot[only marks,draw opacity=0.5,mark=x,blue] table [x=sampled_dist_power
, y=sampled_dist_turbulence, col sep=comma] {figures/mo_floris/4_turbines/data/dist_2.csv};

  \nextgroupplot[xlabel = {power}, grid=major, yticklabel style={/pgf/number format/.cd,fixed,precision=3}]
      \addplot[only marks,draw opacity=0.5,mark=x,red] table [x=learned_dist_power, y=learned_dist_turbulence, col sep=comma] {figures/mo_floris/4_turbines/data/dist_3.csv};
      \addplot[only marks,draw opacity=0.5,mark=x,blue] table [x=sampled_dist_power
, y=sampled_dist_turbulence, col sep=comma] {figures/mo_floris/4_turbines/data/dist_3.csv};

    \nextgroupplot[xlabel = {power}, grid=major, yticklabel style={/pgf/number format/.cd,fixed,precision=3}]
       \addplot[only marks,draw opacity=0.5,mark=x,red] table [x=learned_dist_power, y=learned_dist_turbulence, col sep=comma] {figures/mo_floris/4_turbines/data/dist_4.csv};
      \addplot[only marks,draw opacity=0.5,mark=x,blue] table [x=sampled_dist_power
, y=sampled_dist_turbulence, col sep=comma] {figures/mo_floris/4_turbines/data/dist_4.csv};
    \end{groupplot}
  \end{tikzpicture}
  
  \caption{The learned distributions (learned by real-NVP using DMOVE, red) and the underlying distributions (blue) of power and turbulence for each group in a 4 turbine wind farm.}
    \label{fig:learned_distributions}
\end{figure}
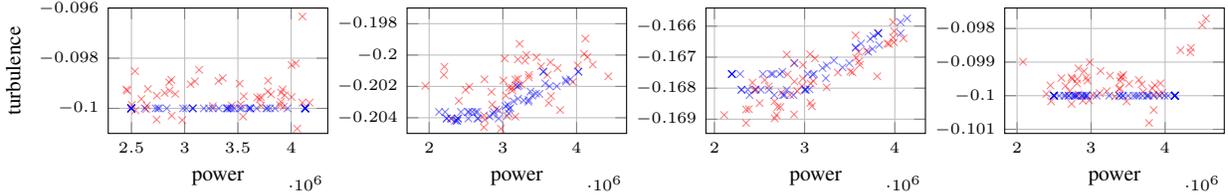

Figure \ref{fig:esr_set} presents the expected value vectors of ESR non-dominated return distributions of the \emph{ESR sets} computed by DMOVE for wind farms with 4, 8, 9 and 10 turbines. Given the large number of policies in the \emph{ESR set} it is not possible to present the distributions for each individual policy. However, Figure \ref{fig:learned_distributions} presents the return distributions for a policy in the \emph{ESR set} computed by DMOVE for each group in a 4 turbine wind farm. Figure \ref{fig:learned_distributions} contrasts the learned distributions (red) with the underlying return distributions (blue). It is clear from  Figure \ref{fig:learned_distributions} the \emph{real-NVP} models utilised by DMOVE can accurately learn the underlying return distributions for each specific turbine.

The learned distributions presented in Figure \ref{fig:learned_distributions} highlight the effectiveness \emph{real-NVP} models have in learning continuous return distributions. To date only categorical distributions have been used to compute policies for the ESR criterion. As highlighted by \citet{hayes2022modvi_esr}, categorical distributions have limitations when dealing with continuous data. Therefore the return distributions learned when using multi-objective categorical methods cannot accurately represent return distributions when the returns are continuous. The results presented in Figure \ref{fig:learned_distributions} show that \emph{real-NVP} models can be used to replace categorical distributions and learn good representations of return distributions in real-world settings with continuous returns.

Given that wind farm control is subjected to strict regulations and constraints, it is crucial that an operator has sufficient information about the potential outcomes that a policy execution may have and the likelihood of each outcome. Therefore, using the distributions presented in Figure \ref{fig:learned_distributions} an operator can perform a more informed analysis of potential policies. Given that in wind farm control regulations and constraints need to be adhered to at any moment, and thus each policy execution is important, having the outlined distributions available can help the operator avoid selecting policies which may have some likelihood of violating these regulations or constraints.

Given the rise in deployment of artificial intelligence systems in the real-world, it is important such systems are explainable, trustworthy and safe \citep{mannion2021trustworthyai}. Distributional methods, like DMOVE, can be used to provide explainability for multi-objective decision making processes. In the context of wind farm control, risk assessment is crucial when developing operation \& maintenance strategies. Having a distribution over the returns available can aid operators in understanding how and why an adverse event occurred as such an event will be present in the distribution of returns. Additionally, distributional methods can also bring outlier events to the attention of the operator before a decision is made, which can help the operator to take decisions to avoid unsafe events. Therefore, it may be possible to utilise distributional multi-objective approaches for expanding the field of trustworthy and safe artificial intelligence \citep{mannion2021trustworthyai}. As already highlighted, the state-of-the-art MORL approaches focuses on computing policies using expected value vectors \citep{roijers15computing, bryce2007probabilistic}. Making decisions based on expected value vectors reduces the explainability and safety of such algorithms given the distribution over the potential outcomes is lost when the expectation in computed. For example, risks in wind farm control depend on short-term events that may occur (e.g., unexpected downtime or grid curtailment). An expected value vector may not effectively capture uncommon events given their probability of occurring may be small. Expected value vectors can also be difficult to interpret especially in environments where outcomes can be stochastic, like wind farm control. In contrast, return distributions can easily be interpreted by a system expert. Therefore, distributional methods could be utilised as a decision making aid in complex real world problem domains.

\section{Related Work}
Multi-objective variable elimination focuses on computing sets of optimal policies for the SER criterion \citep{rollon2006bucket, roijers15computing, marinescu2012multi}. For example, Pareto multi-objective variable elimination (PMOVE) \citep{rollon2006bucket},  and convex multi-objective variable elimination (CMOVE) \citep{roijers15computing} are used to compute the Pareto front and convex coverage set respectively. The outlined methods all utilise expected value vectors to determine a partial ordering over policies and therefore cannot be used to learn optimal policies for the ESR criterion \citep{hayes2022modvi_esr, hayes2022decision}.

The majority of research for the expected scalarised returns criterion largely focuses on utilising categorical distributions to compute return distributions. \citet{hayes2022modvi_esr} compute the \emph{ESR set} using multi-variate categorical distributions in sequential decision making settings when the utility function is unknown. \citet{reymond2021esractor} utilise a multi-variate categorical distribution to learn a return distribution in MORL settings where the utility function of a user is known. However, in this work we utilise generative flow models, which is a class of generative modelling.

Generative modelling \citep{tomczak2022deep} has become a very popular area of research where models such as general-adversarial networks \citep{goodfellow2014generative} (GANs), variational auto-encoders \citep{kingma2013auto} (VAEs) and generative flow models \citep{kingma2018glow,dinh2016density} have made significant advancements in artificial intelligence specifically in the area of supervised learning. Recently, generative flow models have been used in sequential decision making settings to model complex distributions. \citet{ma2020normalizing} utilise a \emph{real-NVP} model conditioned on a state in reinforcement learning settings to model the complex distribution of a policy. 

\section{Conclusion \& Future Work}
In this work, we present a novel \emph{distributional multi-objective variable elimination} (DMOVE) algorithm that can compute the \emph{ESR set} for real-world problem domains where the returns are continuous. We show that DMOVE can compute the \emph{ESR set} in real-world wind farm settings and also highlight how the \emph{real-NVP} models can accurately learn the continuous return distributions by interacting with the environment.

In a future publication we aim to perform a computational analysis on DMOVE to calculate the computational implications of maintaining return distributions to calculate policies. We also aim to utilise generative flow models in multi-objective sequential decision making settings, like multi-objective Markov decision processes (MOMDPs), to compute the \emph{ESR set} in MOMDPs with continuous state-action space and continuous returns.

% Acknowledgements should go at the end, before appendices and references
%\section*{Acknowledgements}
%Conor F. Hayes is funed.

\vskip 0.2in
\bibliographystyle{abbrvnat}
\bibliography{sample}


\newpage

\appendix
\section{Background}
\subsection{Real-NVP}
\begin{figure}[h]
    \centering
     \begin{tikzpicture}[>={Stealth[width=6pt,length=9pt]}, skip/.style={draw=none}, shorten >=1pt, accepting/.style={inner sep=1pt}, auto]
     \draw (0.0pt, 0.0pt)node[regular polygon,regular polygon sides=8, fill=none, thick, minimum height=1.4cm,minimum width=1.4cm, draw](0){};
     \draw (80.0pt, 0.0pt)node[regular polygon,regular polygon sides=8, fill=none, thick, minimum height=0.6cm,minimum width=0.6cm, draw](1){$+$ $\odot$};
     \draw (160.0pt, 0.0pt)node[regular polygon,regular polygon sides=8, fill=none, thick, minimum height=1.4cm,minimum width=1.4cm, draw](2){=};
     \draw (240.0pt, 0.0pt)node[regular polygon,regular polygon sides=8, fill=none, thick, minimum height=0.6cm,minimum width=0.6cm, draw](3){$+$ $\odot$};
     
     \draw (0.0pt, -100.0pt)node[regular polygon,regular polygon sides=8, fill=none, thick, minimum height=1.4cm,minimum width=1.4cm, draw](4){};
     \draw (80.0pt, -100.0pt)node[regular polygon,regular polygon sides=8, fill=none, thick, minimum height=1.4cm,minimum width=1.4cm, draw](5){=};
     \draw (160.0pt, -100.0pt)node[regular polygon,regular polygon sides=8, fill=none, thick, minimum height=0.6cm,minimum width=0.6cm, draw](6){$+$ $\odot$};
     \draw (240.0pt, -100.0pt)node[regular polygon,regular polygon sides=8, fill=none, thick, minimum height=1.4cm,minimum width=1.4cm, draw](7){=};

     %\path[->] (6) edge[dotted, bend right = 50]
     \path[->] (0) edge[] node[above]{} (1);
     \path[->] (1) edge[] node[above]{} (2);
     \path[->] (2) edge[] node[above]{} (3);
     
     \path[->] (4) edge[] node[above]{} (5);
     \path[->] (5) edge[] node[above]{} (6);
     \path[->] (6) edge[] node[above]{} (7);

     \path[->] (4) edge[] node[above]{} (1);
     \path[->] (1) edge[] node[above]{} (6);
     \path[->] (6) edge[] node[above]{} (3);
    
     ;
      
    \end{tikzpicture}
    
    \caption{Alternating pattern for updating \emph{affine coupling layers}.}
    \label{fig:alternating_affine_coupling_layer}
\end{figure}
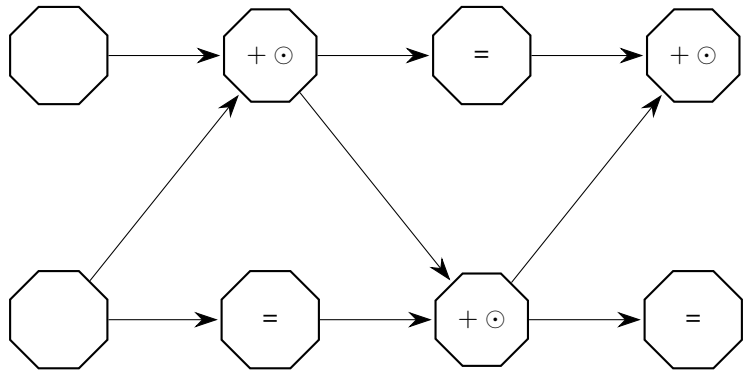

Figure \ref{fig:real_nvp_graph} presents a computational graph for a \emph{real-NVP} model for a single \emph{affine coupling layer}. However, to ensure that all components are changed during training an alternating pattern for multiple \emph{affine coupling layers} must be used for each \emph{real-NVP} model. Figure \ref{fig:alternating_affine_coupling_layer} outlines an alternating patter for multiple \emph{affine coupling layers}, therefore if some components are unchanged in one layer, the components will be changed in the next layer \cite{dinh2016density}.
\label{appendix:background_real_nvp}
\begin{figure}[h]
    \centering
     \begin{tikzpicture}[>={Stealth[width=6pt,length=9pt]}, skip/.style={draw=none}, shorten >=1pt, accepting/.style={inner sep=1pt}, auto]
     \draw (0.0pt, 0.0pt)node[regular polygon,regular polygon sides=8, fill=none, thick, minimum height=0.6cm,minimum width=0.6cm, draw](0){$x_{1}$};
     \draw (60.0pt, 0.0pt)node[circle, thick, minimum height=0.6cm,minimum width=0.6cm, draw](1){$=$};
     \draw (120.0pt, 0.0pt)node[regular polygon,regular polygon sides=8, fill=none, thick, minimum height=0.6cm,minimum width=0.6cm, draw](2){$y_{1}$};
     \draw (0.0pt, -100.0pt)node[regular polygon,regular polygon sides=8, fill=none, thick, minimum height=0.6cm,minimum width=0.6cm, draw](3){$x_{2}$};
     \draw (40.0pt, -100.0pt)node[circle, fill=none, thick, minimum height=0.6cm,minimum width=0.6cm, draw](4){$\odot$};
     \draw (80.0pt, -100.0pt)node[circle, thick, minimum height=0.6cm,minimum width=0.6cm, draw](5){$+$};
     \draw (120.0pt, -100.0pt)node[regular polygon,regular polygon sides=8, fill=none, thick, minimum height=0.6cm,minimum width=0.6cm, draw](6){$y_{2}$};
     \draw (40.0pt, -60.0pt)node[circle, thick, minimum height=0.6cm,minimum width=0.6cm, draw](8){$s$};
     \draw (80.0pt, -40.0pt)node[circle, thick, minimum height=0.6cm,minimum width=0.6cm, draw](9){$t$};
     
     %\path[->] (6) edge[dotted, bend right = 50]
     \path[->] (0) edge[] node[above]{} (1);
     \path[->] (1) edge[] node[above]{} (2);
     \path[->] (8) edge[] node[above]{} (4);
     \path[->] (9) edge[] node[above]{} (5);
     \path[->] (3) edge[] node[above]{} (4);
     \path[->] (4) edge[] node[above]{} (5);
     \path[->] (5) edge[] node[above]{} (6);
     \path[->] (0) edge[] node[above]{} (8);
     \path[->] (0) edge[] node[above]{} (9);
     ;
      
    \end{tikzpicture}
    
    \caption{Computational graph for the forward propagation of a \emph{real-NVP} instance.}
    \label{fig:real_nvp_graph}
\end{figure}
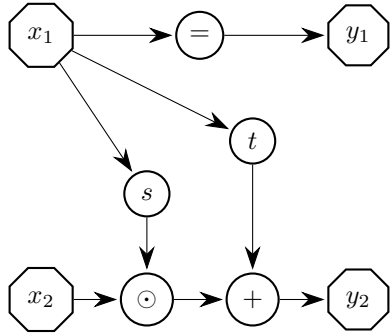

\section{DMOVE}
\subsection{Coordination Graph}
\begin{figure}[h]
    \centering
     \begin{tikzpicture}[>={Stealth[width=6pt,length=9pt]}, skip/.style={draw=none}, shorten >=1pt, accepting/.style={inner sep=1pt}, auto]
     \draw (80.0pt, 0.0pt)node[regular polygon,regular polygon sides=8, fill=none, thick, minimum height=1.4cm,minimum width=1.4cm, draw](1){$f_{i}$};
     \draw (240.0pt, 0.0pt)node[regular polygon,regular polygon sides=8, fill=none, thick, minimum height=1.4cm,minimum width=1.4cm, draw](3){$f_{i}$};
     
     \draw (0.0pt, -100.0pt)node[regular polygon,regular polygon sides=8, fill=none, thick, minimum height=1.4cm,minimum width=1.4cm, draw](4){$n_{i}$};
     \draw (160.0pt, -100.0pt)node[regular polygon,regular polygon sides=8, fill=none, thick, minimum height=1.4cm,minimum width=1.4cm, draw](6){$i$};
     
     \path[->] (4) edge[] node[above]{} (1);
     \path[->] (6) edge[] node[above]{} (1);
     \path[->] (6) edge[] node[above]{} (3);
    
     ;
      
    \end{tikzpicture}
    
    \caption{Coordination graph showing the graphical relationship between agent $i$ and neighbouring RSFs $f_{i}$ and neighbouring agents $n_{i}$.}
    \label{fig:coordination_graph_i}
\end{figure}
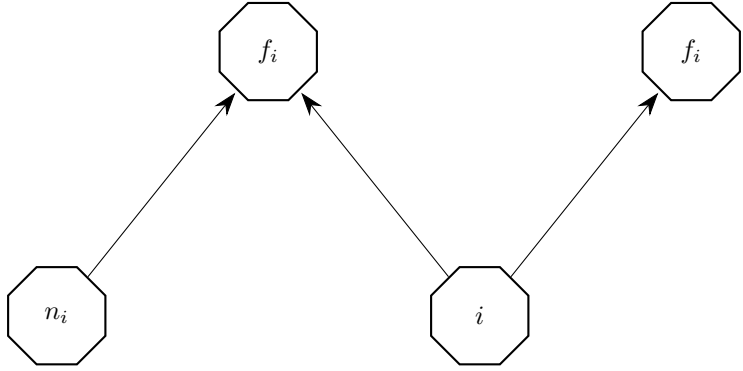

Figure \ref{fig:coordination_graph_i} outlines for agent $i$ the neighbour nodes, $n_{i}$ and the set of RSFs, $f_{i}$, which agent $i$ influences.
\label{sec:dmove_coordination_graph}

\subsection{ESR Prune}
Algorithm \ref{alg:ESRPrune} outlines the \texttt{ESRPrune} algorithm \citep{hayes2022decision} which is used as \texttt{prune1}, \texttt{prune2} and \texttt{prune3} in DMOVE.
\label{appendix:esr_prune}
\begin{algorithm}[h]
\SetAlgoLined
\textbf{Input}: $\textbf{Z}\ \leftarrow$ \text{A set of return distributions}\\
$\textbf{Z}^{*}\ \leftarrow \emptyset$ \\
\While{$\textbf{Z} \neq \emptyset$}
{$\textbf{z}  \leftarrow \text{the first element of $\textbf{Z}$}$ \\
\For{$\textbf{z}' \in \textbf{Z}$}{\If{$\textbf{z}' >_{ESR} \textbf{z}$}{${\bf{z}} \leftarrow \textbf{z}'$}}\text{Remove $\textbf{z}$ and all return distributions} \text{ESR-dominated by $\textbf{z}$ from $\textbf{Z}$}\\ \text{Add $\textbf{z}$ to $\textbf{Z}^{*}$} }
\textbf{Return $\textbf{Z}^{*}$}
 \caption{\texttt{ESRPrune}}
 \label{alg:ESRPrune}
\end{algorithm}

\section{Experiments}
\subsection{Real-NVP}
\label{appendix:real_nvp}
Table \ref{tab:real_nvp-params} outlines the parameters used by DMOVE, each \emph{real-NVP} and the $\texttt{learn}$ algorithm for each wind farm used during experimentation. The parameter $flows$ determines the number of \emph{affine coupling layers} used in each \emph{real-NVP} model. While the optimiser and lr parameters determine which optimiser is used and what the learning rate is set to for each \emph{real-NVP} model. To compute the scale, s, and transformation, t, functions of a \emph{real-NVP} model we represent both s and t with a neural network where the hidden layers parameter determines the number of hidden layers in the network. We note that $r_{max}$, $r_{min}$, $n_{samples}$ and $n_{bins}$ are used to compute the cumulative distribution function (CDF) of each return distribution. The $n$ parameter determines the number of samples which are taken from the replay buffer during training, the size of the replay buffer size is determined by the replay buffer size parameter. Samples from the environment are gathered by executing a random action a number of times determined by the parameter steps. Finally, we train each \emph{real-NVP} model in increments, $t_{inc}$. Therefore training takes places every $t_{inc}$ steps.
\begin{table*}
    \centering
    \begin{tabular}{|c|c|c|c|c|}
        \hline
         & \textbf{4 Turbines} & \textbf{8 Turbines} & \textbf{9 Turbines} &  \textbf{10 Turbines} \\
        \hline
        \hline
        $flows$ & $8$ & $8$ & $8$ & $8$ \\
        $optimiser$ & Adam & Adam & Adam & Adam \\
        $lr$ & $1e^{-3}$ & $1e^{-3}$ & $1e^{-3}$ & $1e^{-3}$ \\
        $hidden$ $layers$ & $30$ & $30$ & $30$ & $30$ \\
        $\textbf{r}_{min}$ & [$-1.0, 0.0$] & [$-2.0, 0.0$] & [$-2.0, 0.0$] & [$-2.0, 0.0$] \\
        $\textbf{r}_{max}$ & [$0.0, 2.0e^{7}$] & [$0.0, 3.5e^{7}$] & [$0.0, 4.0e^{7}$] & [$0.0, 4.2e^{7}$] \\
        $n_{bins}$ & $2,000$ & $2,000$ & $2,000$ & $2,000$ \\
         n & $500$ & $500$ & $500$ & $500$ \\
         $n_{samples}$ & $2,000$ & $2,000$ & $2,000$ & $2,000$ \\
        $t_{inc}$ & $1,000$ & $1,000$ & $1,000$ & $1,000$ \\
        replay buffer size & $5,000,000$ & $5,000,000$ & $5,000,000$ & $5,000,000$ \\
        steps & $300,000$ & $300,000$ & $300,000$ & $300,000$ \\

        \hline
       
    \end{tabular}
    \caption{Hyperparameters for the experiments for DMOVE and each corresponding wind farm.}
    \label{tab:real_nvp-params}
\end{table*}

\section{Related Work on Wind Farm Control}
Wind farm control research mainly focuses on the maximization of generated power and minimization of fatigue loads (i.e., stress induced on the mechanical components during steady-state operation) \citep{tutorial_control_2017}. Data-driven optimization approaches are often used to optimize control parameters in a wake simulator \citep{verstraeten2021scalable}. This is generally achieved by reducing the wake effect. In our work, we focus on wake redirection control, which aims at finding a joint orientation of the wind turbines' nacelles to redirect wake away from the wind farm \citep{vandijk2016,wagenaar2012controlling}. State-of-the-art methods focus on performing analysis of wake redirection strategies on small-sized wind farms. Moreover, although both power and fatigue load are considered as optimization criteria, a simple linear scalarization of the objectives is typically used \citep{vandijk2016}. In our work, we demonstrate the benefits of using our a multi-objective approach in the context of wake redirection control to provide a more flexible decision framework.

%\subsection{Local Distributions}

%\begin{figure}
%    \centering
%    \plotPolicies{1}{20}
%    \caption{Execution of policies 1 to 20.}
%    \label{fig:policy-executions-ode}
%\end{figure}

%\begin{figure}
%    \centering
%    \plotPolicies{21}{40}
%    \caption{Execution of policies 1 to 20.}
%   \label{fig:policy-executions-ode}
%\end{figure}

%\begin{figure}
%    \centering
%    \plotPolicies{41}{60}
%    \caption{Execution of policies 1 to 20.}
%    \label{fig:policy-executions-ode}
%\end{figure}

%\begin{figure}
%    \centering
%    \plotPolicies{61}{80}
%    \caption{Execution of policies 1 to 20.}
%    \label{fig:policy-executions-ode}
%\end{figure}

\end{document}